\pdfoutput=1

\documentclass[11pt]{article}
\usepackage{ulem}

\usepackage[final]{acl}

\usepackage{times}
\usepackage{latexsym}
\usepackage{listings}
\usepackage{pdfpages}
\usepackage{bm}
\usepackage{array}
\usepackage{makecell}
\usepackage{tabularx}
\usepackage[T1]{fontenc}

\usepackage[utf8]{inputenc}
\usepackage{microtype}
\usepackage{inconsolata}
\usepackage{graphicx}
\usepackage{multirow}
\usepackage{booktabs}
\usepackage{amsmath}
\usepackage{ulem}
\usepackage{balance}

\newcommand{\paratitle}[1]{\vspace{0.2cm}\noindent\textbf{#1}}

\title{BOSE: A Systematic Evaluation Method Optimized for Base Models}

\author{
Hongzhi Luan\footnotemark[1], \ Changxin Tian\thanks{
 Equal contributions.}, \ Zhaoxin Huan \\
\textbf{\ Xiaolu Zhang, \ Kunlong Chen, \ Zhiqiang Zhang, \ Jun Zhou\thanks{Corresponding author.}} \\
Ant Group \\
  \texttt{\{luanhongzhi.lhz, tianchangxin.tcx, zhaoxin.hzx\}@antgroup.com} \\
  \texttt{\{yueyin.zxl, kunlong.ckl, lingyao.zzq, jun.zhoujun\}@antgroup.com}
\footnotetext[1]{Equal contributions.}
}

\begin{document}
\maketitle
\begin{abstract}
This paper poses two critical issues in evaluating base models (without post-training): (1) Unstable evaluation during training: in the early stages of pre-training, the models lack the capability to answer questions as required, leading to unstable evaluation results. This instability makes it difficult to provide solid conclusions to guide the training, especially for key experiments such as data ablation and scaling law. 
(2) Inconsistency between base and instruct models: base models generally exhibit poorer evaluation performance compared to corresponding instruct models. This gap poses a challenge for assessing whether a base model with better evaluation can truly lead to a better instruct model. To address these issues, we propose \textbf{B}ase model \textbf{O}riented \textbf{S}ystematic \textbf{E}valuation (\textbf{BOSE}), a method specifically designed to optimize the evaluation of base models. Specifically, BOSE introduces two key innovations: In-Context Light-instruction Prompt (ICLiP) for open-ended tasks and Blank-ppl for multi-choice tasks with candidate options, which transforms the standard perplexity (ppl) metric into a fill-in-the-blank format to mitigate early-stage evaluation fluctuations. Furthermore, we are the first to propose Kendall's rank correlation to quantitatively measure the evaluation stability and consistency. Experimental results demonstrate that BOSE significantly enhances both the stability of evaluations during pre-training and the consistency between base and instruct models, thereby providing more reliable guidance for the LLMs' training.
\end{abstract}

\section{Introduction}

Recently, large language models (LLMs) have demonstrated remarkable achievements across various domains~\cite{llm-survey}. This has led to the development of numerous high-performing LLMs~\cite{gpt4,llama3,qwen2.5,gemma2}. To evaluate the performance of these models on a wide range of tasks, an increasing number of benchmarks have been open-sourced~\cite{llmevaluation-survey}. These benchmarks provide a comprehensive evaluation of the capabilities of LLMs, guiding the training and improving their weakness. Evaluating large language models is a cornerstone in development of LLMs.

Generally, LLMs can be divided into two primary categories based on whether they receive post-training: \textit{base models} and \textit{instruct models}~\cite{llama3, fu2022gptroadmap}. 
Instruct models undergo downstream adaptation, allowing them to adapt to specific tasks and answer questions as required. In contrast, base models focus on learning foundational knowledge without targeting specific tasks, resulting in a weaker ability to respond to instructions. Consequently, most current benchmarks provide comprehensive evaluations of instruct models~\cite{Multi_LogiEval,MGSM}, while evaluations of base models are relatively rare. Nevertheless, base models serve as the foundation for the entire LLM training. An accurate evaluation for base models can provide essential insights for training LLMs, such as experiments on scaling laws~\cite{scalinglaw}, ablation of pre-training data, and the selection of appropriate base checkpoints for post-training~\cite{llama3, deepseekv3}.

\begin{figure*}[ht]
\centering
  \includegraphics[width=\textwidth]{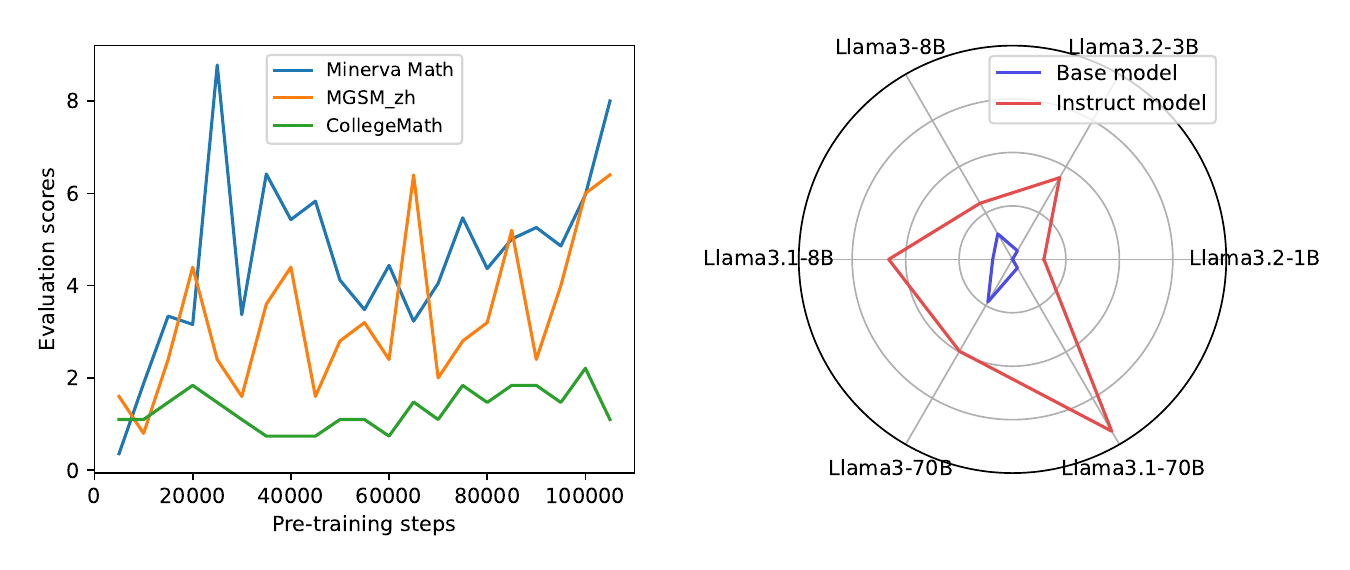}
  \caption{Illustrations of the critical issues in current base model evaluation. (\textbf{left}): Evaluation scores of 3 benchmarks on a series of pre-trained checkpoints, with the x-axis representing the increasing trained steps, and the y-axis representing the evaluation scores. The evaluation scores do not improve stably during pre-training. (\textbf{right}): Comparison of evaluation scores on 6 Llama base models and instruct models for several benchmarks. There is a lack of consistency between the two score series. For example, Llama-3.1-70B achieves the highest score among instruct models, but performs poorly in the base model.}
  \label{fig:intro_figure}
\end{figure*}



In this paper, we argue that the current evaluation is not suited to the characteristics of base models and lacks a systematic methodology. In a nutshell, the challenge arises from two dimensions:

\begin{itemize}
    \item \textbf{Instability of evaluations during training}. Critical pre-training experiments, such as data ablation and scaling laws, rely on the performance of base models in the early stages of training. However, base models cannot follow instructions and solve problems during the initial stages, leading to fluctuations in evaluation results. Figure~\ref{fig:intro_figure} (left) illustrates the dynamic evolution of a 1B-parameter model's performance across different benchmarks as the training tokens increase. The significant fluctuations in the evaluation make it difficult to determine whether the model has gained real improvement throughout the training tokens. This instability hinders making reliable decisions for these critical experiments.

    \item \textbf{Lack of consistency in evaluating base and instruct models}. During pre-training, LLMs acquire more knowledge than the post-training. However, base models typically perform worse on evaluations than their corresponding instruct models due to the weaker instruction-following capabilities. Taking the Llama family models~\cite{llama3} as an example, we evaluate these models on several recently released mathematical datasets~\cite{CollegeMath,MinervaMath,MGSM}, as shown in Figure~\ref{fig:intro_figure} (right). The base models' scores are significantly lower than their corresponding instruct models, and there is little distinction among the different versions of the base models. This discrepancy presents a challenge for current evaluation methods in determining whether a base model with better evaluation results can indeed lead to a better instruct model.
    
\end{itemize}

In this paper, we propose a systematic approach for the evaluation of base models, called \textbf{B}ase model-\textbf{O}riented \textbf{S}ystematic \textbf{E}valuation (\textbf{BOSE}). BOSE aims to address the aforementioned critical issues by optimizing existing evaluation methods.
Specifically, we observe that base models often struggle to understand evaluation questions, which results in inaccurate reflections of their true capabilities. This is particularly evident in their difficulty with multi-choice questions, poor ability to follow complex instructions, and tendency to generate redundant continuations. BOSE improves evaluation techniques tailored to pre-training characteristics for two key tasks: open-ended generation and multi-choice. It introduces \textbf{I}n-\textbf{C}ontext \textbf{L}ight-\textbf{i}nstruction \textbf{P}rompt (\textbf{ICLiP}) to improve the base model's responses to open-ended questions and transforms multi-choice questions into fill-in-the-blank versions (\textbf{Blank-ppl}) to mitigate evaluation fluctuations in early stage. Finally, we validate the effectiveness of BOSE by examining two key aspects: training stability and consistency between base and instruct models. This evaluation aims to determine whether the results exhibit a stable growth trend and whether a better base model can lead to a better instruction model.




In summary, our contributions are as follows:
\begin{itemize}
    \item We propose BOSE, a systematic evaluation method tailored to base models that incorporates ICLiP for open-ended generation tasks and blank-ppl for multi-choice tasks. 
    This approach aims to better align with the characteristics of base models, yielding evaluation results that accurately reflect their true capabilities.
    \item To assess the effectiveness of base model evaluation method, we define criteria based on the stability of metrics during pre-training and the consistency of capabilities between base and corresponding instruct models, and pioneer the use of Kendall's rank correlation as a quantitative metric.
    \item We conduct comprehensive experiments on multiple benchmarks covering knowledge, mathematics and reasoning, using both open-source models and our pre-trained checkpoints. Empirical results demonstrate that BOSE significantly improves the evaluation stability during pre-training and enhances the consistency between the base and instruct models, which is highly beneficial for guiding the model development and ensuring the reliability of base model evaluation results.
\end{itemize}
To the best of our knowledge, this paper presents the first systematic framework for evaluating base LLMs, introducing empirically validated optimization methodologies with quantitative metrics. We believe this will yield actionable insights and practical recommendations to contribute to advance future base LLM evaluation. 

\section{Related Work}
In this section, we first introduce some commonly used benchmarks and evaluation tasks for LLMs, and then focus on the existing approaches for evaluating base models.

\paratitle{Evaluation Tasks.} 
Based on whether there exist reference answers to be automatically calculated, evaluation tasks commonly can be classified as either ground truth-based evaluation or human preference-based evaluation~\cite{chatbot-arena, llmevaluation-survey}.
Considering that base models typically cannot align with human preferences, we focus on evaluation tasks with ground truths. As for these benchmarks, the presence of candidate answers can generally further differentiate~\cite{2023opencompass}:
  \begin{itemize}
      \item \textbf{Open-ended task}. This type of evaluation tasks require the model to respond to questions according to given instructions, and employ a customized post-processing process to extract potential answers as well as judge the correctness, as seen in benchmarks such as Math~\cite{math}, BBH~\cite{bbh} and HumanEval~\cite{humaneval}.
      \item \textbf{Multi-choice task}. In these tasks, given a question, the model needs to choose the most appropriate option from multiple choices. As for evaluating base models, perplexity for a given sentence is typically calculated to evaluate a model’s language modeling capabilities~\cite{gpt3}, as seen in benchmarks such as MMLU~\cite{mmlu}, CMMLU~\cite{cmmlu}, etc.
  \end{itemize}


\paratitle{Base Model Evaluation.}
To our knowledge, the evaluation of base models remains under-addressed. Despite the availability of numerous benchmarks, few provide detailed evaluations of base models. For instance, CMath~\cite{CMath} releases datasets but pays little attention to evaluation prompts. Many benchmarks, such as SimpleQA~\cite{simpleQA}, MuSR~\cite{MuSR}, Multi\_LogiEval~\cite{Multi_LogiEval}, and CollegeMath~\cite{CollegeMath}, do not distinguish between the evaluation of base and instruct models. Only a few benchmarks, like kor-bench~\cite{kor-bench}, consider both aspects.

A common approach to evaluate base models is using in-context learning~\cite{gpt3, ICL-survey}, particularly the few-shot method. This method expects the base model to leverage its in-context learning ability to respond to questions based on given examples, thereby addressing instruction-following challenges. Therefore, some open-source evaluation frameworks, such as lm-eval~\cite{lm-eval-harness} and OpenCompass~\cite{2023opencompass} utilize in-context learning to evaluate the performance of base models. However, this approach lacks systematic guidance on how to effectively conduct in-context learning, especially when considering the characteristics of base models in different training stages.


Due to the above issues, although many technical reports of open-source LLMs disclose the benchmarks and evaluation results used for base model evaluation~\cite{llama3, qwen2.5, gemma2}, the lack of unified evaluation protocol and the incomplete transparency of evaluation details make it challenging to reproduce and align the evaluation results, which may cause significant difficulties for researchers. 

\section{Methodology}
Different from instruct models, base model evaluation encounters unique challenges due to the lack of instruction-tuning and preference alignment. Specifically, base models typically fail to comprehend the intent behind prompts or follow explicit instructions, leading to suboptimal evaluation protocols that neither reflect their intrinsic capabilities nor bring reliable evaluation results, which becomes particularly critical during early pre-training stages.

\begin{figure}[tbp]
  \includegraphics[width=\columnwidth]{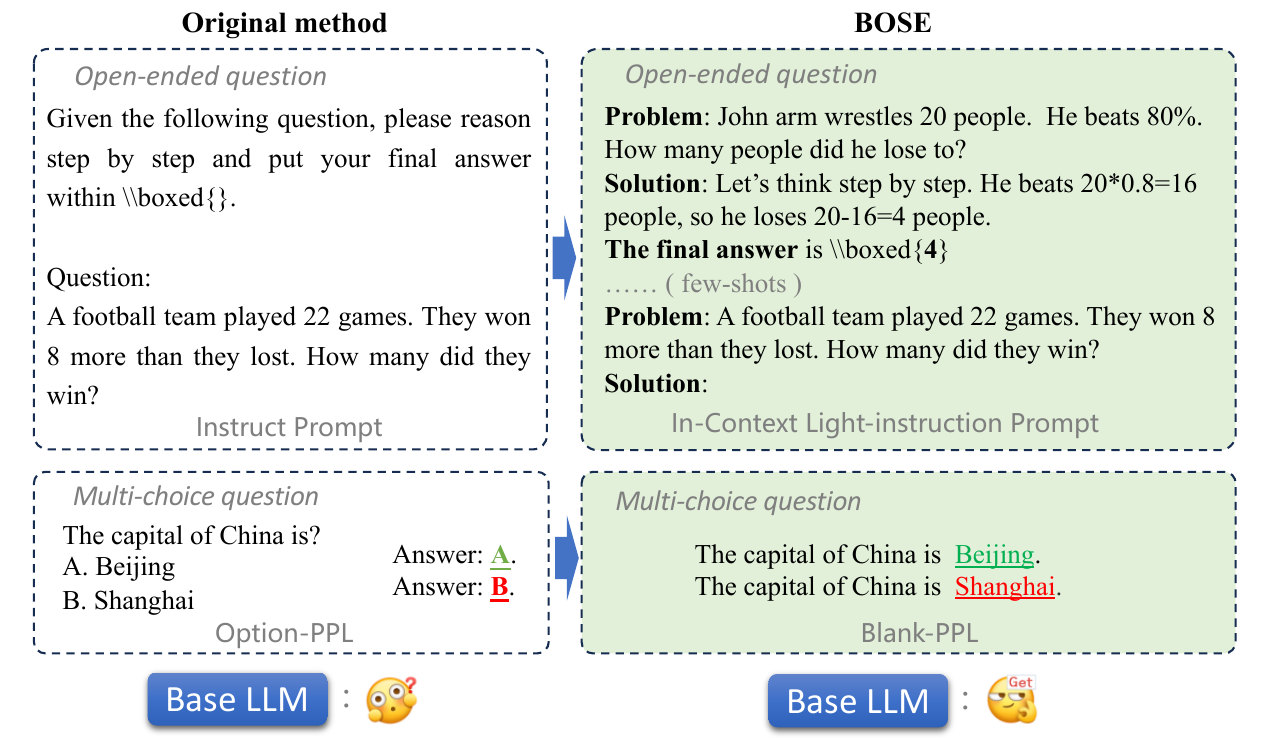}
  \caption{Illustrations of our proposed BOSE, which incorporates ICLiP for open-ended generation tasks (\textbf{top}) and Blank-ppl for multi-choice tasks (\textbf{bottom}).}
  \label{fig:method}
\end{figure}

To facilitate these limitations and align evaluation protocols with the intrinsic characteristics of base models, we optimize the evaluation methods for open-ended and multi-choice tasks respectively, as illustrated in Figure~\ref{fig:method}.

\subsection{Open-ended Task}
Considering the inherent characteristics of base models in generative tasks, we propose the In-Context Light-instruction Prompt (\textbf{ICLiP}), which comprises three core components:

\textbf{In-context learning}. In-context learning (ICL)~\cite{gpt3, ICL-survey} is an effective paradigm that enables pre-trained language models to perform new tasks without gradient updates. By providing some carefully prepared examples in the form of demonstrations, the model implicitly learns to mimic the reasoning patterns and output styles demonstrated in the context, thereby exhibiting its intrinsic problem-solving abilities. Our ICLiP method incorporates standardized few-shot examples within the in-context learning paradigm.

\textbf{Light-instruction prompt}. While prompting techniques are effective for LLMs~\cite{prompt-survey, cot}, excessive prompting may confuse base models due to their lack of instruction-tuning~\cite{cot_decoding}. 
We propose a lightweight prompt template, which formulates the input using the following format: 
\begin{flushleft}\small
\texttt{
\textbf{Problem}: \{problem\} } \\ 
\texttt{
\textbf{Solution}: let's think step by step. \{cot\}} \\ 
\texttt{
\textbf{The final answer is} $\backslash$boxed{\{answer\}}
}
\end{flushleft}
where ``\texttt{\{problem\}}'' represents the problem to be solved, ``\texttt{\{cot\}}'' demonstrates the intermediate reasoning steps, 
and the final answer is presented in a specified format for few-shot examples. In the target question, ``\texttt{\{cot\}}'' is left empty.
We refer to this prompt template as \textbf{light-instruction}, as shown in Figure~\ref{fig:method}, demonstrating better adaptation to the pre-training paradigm and intrinsic abilities of base models.


\textbf{With stopping criteria}. Using the above few-shot light-instruction prompt, a possible consequent issue is the uncontrolled continuations in generation, in other words, base model may fail to complete the answering and continue to generate another question instead. This not only degrades evaluation efficiency but also complicates the optimal answer extraction (e.g., retrieving the last numerical value as the final answer). To mitigate this, we augment the EOS token list in the \verb|generate| function with a special text "\verb|Problem:|". When the model encounters these tokens during decoding, it terminates response immediately and prevents redundant outputs.

\subsection{Multi-choice Task}
Standard evaluation of multi-choice tasks typically employs perplexity-based method to derive the optimal option, and we refer to it as option-ppl in this paper. However, we encounter critical limitations in this approach when evaluating base models with smaller parameter sizes or those in early pre-training stages. 

The dominant approach to pre-training LLMs typically involves language modeling, which is commonly framed as a next-token-prediction task~\cite{GPT, llama3}. 
To better align with the inherent architecture of this pre-training process, we reformulate the option-ppl as a fill-in-the-blank format, leveraging the natural sequential structure inherent in pre-training corpora. Specifically, as illustrated in Figure~\ref{fig:method}, we omit the candidate options and calculate perplexity for each option with the concatenation of the question and candidate text directly. This ensures stronger contextual coherence for the true answer while presenting higher perplexities for others. 

\section{Experiments}
In this section, we describe systematic experiments to investigate how BOSE enhances evaluation stability during pre-training and ensures consistency between base and instruct models\footnote{All of our experiments are implemented based on OpenCompass~\cite{2023opencompass}, and the prompt templates are available in Appendix \ref{sec:appendixB}.}. 




\subsection{Setup}
\subsubsection{Benchmarks}
We employ 9 benchmarks, mainly categorized into two classes by task type:

\textbf{Open-ended tasks}. We take 5 mathematical reasoning benchmarks: CMath~\cite{CMath}, MGSM\footnote{While the MGSM benchmark provides multilingual variants, we employ the Chinese subset for experimental in this study, and named it with MGSM\_zh for simplicity.}~\cite{MGSM}, Gaokao2023EN~\cite{Gaokao2023EN}, CollegeMath~\cite{CollegeMath}, and Minerva Math~\cite{MinervaMath}, alongside 1 multi-step logical reasoning benchmark Multi\_LogiEval~\cite{Multi_LogiEval}. These benchmarks generally require multiple intermediate thinking steps to derive the final answers, and are primarily employed to validate the proposed ICLiP protocol. Answer extraction and judge functions are implemented before accuracy calculation, with greedy decoding applied throughout all experiments.

\textbf{Multi-choice tasks}. We use 3 knowledge-driven benchmarks (MMLU~\cite{mmlu}, CMMLU~\cite{cmmlu}, and MMLU\_Pro~\cite{mmlu_pro}), specifically to test the blank-ppl methodology. These benchmarks assess model performance through perplexity-based evaluation, where the option with the lowest perplexity is identified as the predicted answer.

We utilize accuracy as the unified evaluation metric for all benchmarks. Notably, these benchmarks are mainly chosen from recent public researches, and encompass both English and Chinese benchmarks across the two categories of tasks, ensuring linguistic diversity and comprehensive experimental validation. It is also applicable to a wider range of benchmarks.

\subsubsection{Models}\label{tb:Models}
To comprehensively evaluate BOSE, our experimental framework incorporates two categories of base models:

\textbf{Pre-trained models}\footnote{Since the open-source LLMs do not provide publicly accessible intermediate model weights, we take our pre-trained models for experiments.}. These models are collected from our pre-training experiments, encompassing 1B and 2B parameter sizes with different training steps, 96 checkpoints in total, more details about the training recipe can be found in Appendix \ref{sec:arch} and our technical report~\cite{ling2025}.
To assess the capability consistency between base models and their instruction-tuned counterparts, we further fine-tune 24 checkpoints using identical SFT data (size, quality, filtering criteria) and training hyper parameters to derive the instruct models.

\textbf{Open-source base models}. These models mainly cover three remarkable LLM families: (1) \textbf{Llama family}~\cite{llama3}: mainly including Llama-3.1-8B/70B, with an extended analysis incorporating Llama-3-8B/70B and Llama-3.2-1B/3B for consistency investigation; (2) \textbf{Gemma family}~\cite{gemma2}: Gemma-2-9B/27B; (3) \textbf{Qwen family}~\cite{qwen2.5}: Qwen2.5-7B/72B.

\begin{table*}[htp]
\centering
\resizebox{0.88\textwidth}{!}{
\begin{tabular}{c|c|ccc|ccc} \hline
\multirow{2}{*}{Tasks}            & \multirow{2}{*}{Benchmark} & \multicolumn{3}{c|}{1B}                                                            & \multicolumn{3}{c}{2B}                                                            \\ 
                                  &                            & Original                  & BOSE                       & \textbf{Improve}          & Original                  & BOSE                       & \textbf{Improve}          \\ \hline
\multirow{6}{*}{Open-ended} & CMath                      & 0.418                     & 0.669                     & \textbf{0.251}            & 0.524                     & 0.736                     & \textbf{0.212}            \\
                                  & MGSM\_zh                   & 0.375                     & 0.234                     & \textbf{-0.141}           & 0.376                     & 0.617                     & \textbf{0.241}            \\
                                  & Gaokao2023EN               & 0.537                     & 0.629                     & \textbf{0.092}            & 0.543                     & 0.597                     & \textbf{0.054}            \\
                                  & CollegeMath              & 0.345                     & 0.684                     & \textbf{0.339}            & 0.691                     & 0.721                     & \textbf{0.030}            \\
                                  & Minerva Math              & 0.264                     & 0.089                     & \textbf{-0.175}           & 0.296                     & 0.200                     & \textbf{-0.096}           \\
                                  & Multi\_LogiEval            & 0.099                     & 0.146                     & \textbf{0.047}            & 0.324                     & 0.362                     & \textbf{0.038}            \\ \hline
\multirow{3}{*}{Multi-choice}              & MMLU                       & 0.637                     & 0.845                     & \textbf{0.208}            & 0.795                     & 0.834                     & \textbf{0.039}            \\
                                  & CMMLU                      & 0.754                     & 0.895                     & \textbf{0.141}            & 0.941                     & 0.893                     & \textbf{-0.048}           \\
                                  & MMLU\_Pro                  & 0.382                     & 0.778                     & \textbf{0.396}            & 0.603                     & 0.874                     & \textbf{0.271}            \\ \hline
\multicolumn{2}{c|}{AVG}                                        & \multicolumn{1}{c}{0.423} & \multicolumn{1}{c}{0.552} & \multicolumn{1}{c|}{\textbf{0.129}} & \multicolumn{1}{c}{0.566} & \multicolumn{1}{c}{0.648} & \multicolumn{1}{c}{\textbf{0.082}} \\ \hline
\end{tabular}
}
\caption{Details of evaluation stabilities on our pre-trained 1B and 2B parameter models}
\label{tb:stability}
\end{table*}


\subsubsection{Metric}
To assess the effectiveness of BOSE, we introduce Kendall's rank correlation \cite{kendall} as a quantitative metric beyond intuitive results. Formally, the Kendall's rank correlation $\tau$ is defined as:
\begin{equation}
\begin{aligned}
    \tau = \frac{P - (n(n-1)/2 - P)}{n(n-1)/2} = \frac{4P}{n(n-1)} - 1 \nonumber
\end{aligned}
\end{equation}
where $n$ denotes the total number of entities, $P$ denotes the count of concordant pairs where two entities maintain identical ranking orders, and  $n(n-1)/2$ represents the total number of possible pairwise comparisons. The correlation ranges between -1 and 1, with 1 indicating perfect concordance, -1 denoting complete discordance, and 0 corresponding to random ordinal association.

Upon this metric, we further introduce two measurements to inspect the effectiveness of base model evaluation results:

\textbf{Stability} during pre-training: calculated with the Kendall's rank correlation between evaluation scores and pre-training tokens sequences. 

\textbf{Consistency} with instruct models: measured by the Kendall's rank correlation between base models' evaluation scores and corresponding instruct models' scores, which is inspired by previous work within the community~\cite{manyshoticl}.

\begin{figure}[tbp]
  \includegraphics[width=\columnwidth]{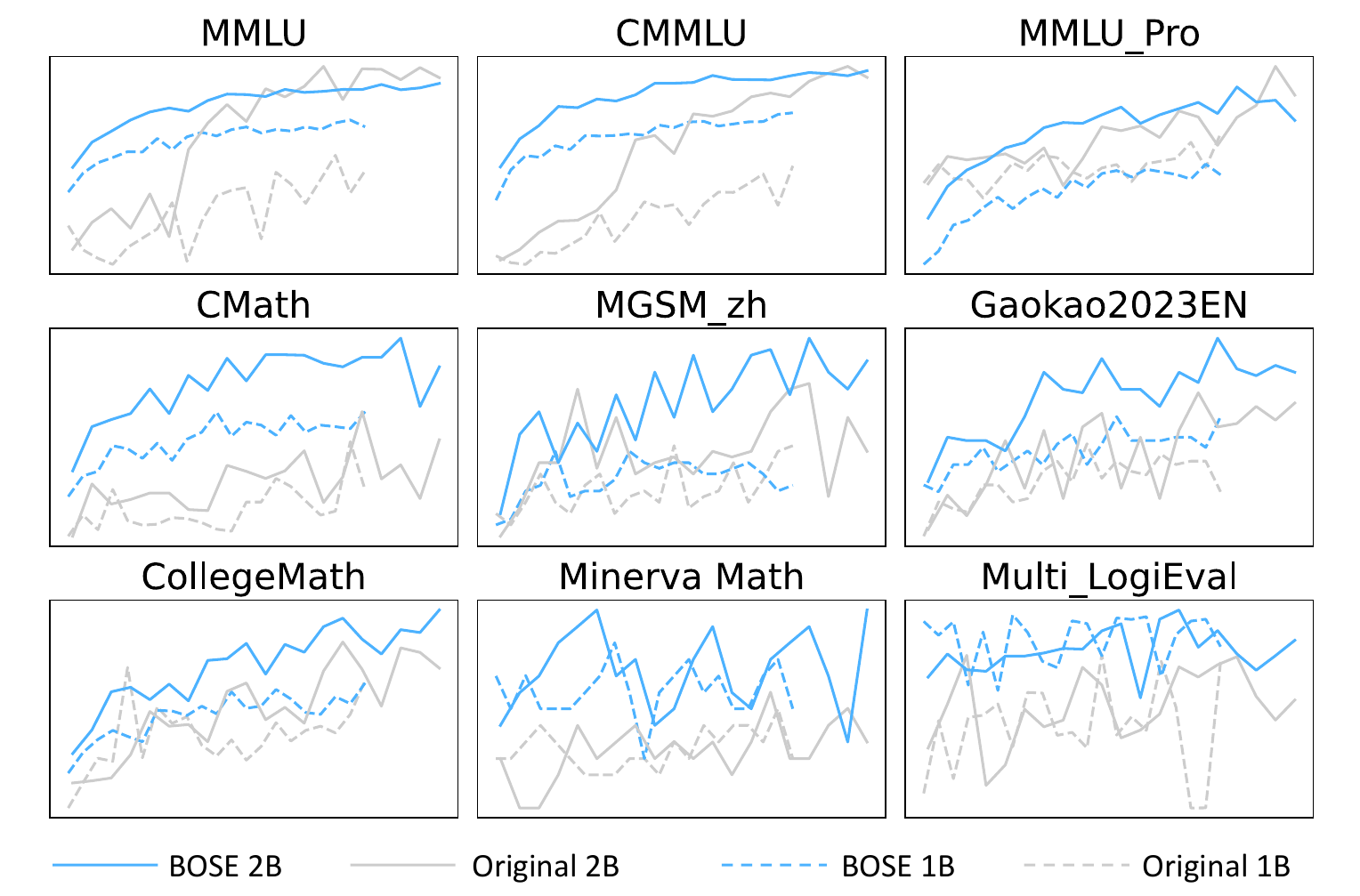}
  \caption{Intuitive results of evaluation stabilities across different benchmarks. X-axis: pre-training steps; y-axis: evaluation scores.}
  \label{fig:model_capability}
\end{figure}

\subsection{BOSE Improves Evaluation Stability During Training}
In this subsection, we validate the stability of model capabilities during pre-training using our pre-trained 1B and 2B parameter models with varying trained steps, including their performance on 6 open-ended tasks and 3 multi-choice tasks.
\subsubsection{Intuitive Results}
As visualized in Figure \ref{fig:model_capability}, our experimental results demonstrate how model evaluation scores grow with increasing trained steps across different benchmarks. We observe that, original evaluation methods exhibit some fluctuations approaching random variation across most benchmarks, whereas BOSE achieves relatively smoother trends. 

We argue that instabilities still emerge in specific benchmarks (e.g., Minerva Math and Multi\_LogiEval), potentially due to the high difficulty of these benchmarks and the limitations in model size or consumed token number.

\subsubsection{Quantitative Metrics}
To further analyze the improvements of evaluation stabilities with BOSE, we calculate Kendall's rank correlations between evaluation scores of two methods and pre-training steps, as detailed in Table \ref{tb:stability}. 

In open-ended tasks, we achieve average $\tau$ improvements of $0.129$ ($+30.5\%$) and $0.082$ ($+14.5\%$) in 1B and 2B parameter models respectively, sustained improvement trends across most tasks except Minerva Math; as for multi-choice tasks, 5/6 experiments demonstrate improved $\tau$ values, with correlation levels remaining high ($\tau = 0.893$ in the single degradation case),  which confirm BOSE significantly enhances evaluation stability during pre-training.

In summary, we demonstrate that BOSE aids in enhancing capability monitoring throughout LLMs’ pre-training, thus providing solid guidance for the training process or early detection of training anomalies through stability metric deviations.


\subsection{ICLiP Enhances the Consistency between Base and Instruct Models}
As discussed previously, we expect to not only ensure stable model evaluation with increasing pre-training tokens but also enhance the consistency between base models and corresponding instruct models, thereby reflecting real model capabilities. This implies a fundamental hypothesis: a base model that performs better in evaluation leads to a better instruct model, and vice versa.

We conduct systematic experiments to compare base models with their post-trained instruct models across both open-source models and our pre-trained checkpoints.
Specifically, for each benchmark, we calculate the Kendall's rank correlation $\tau$ between evaluation results from a series of base models and corresponding post-trained instruct models, of which instruct models are assessed with commonly used instruct prompt, while base models are evaluated using both instruct prompt and our proposed ICLiP method, refer to appendix \ref{sec:appendixA} for further details.

\begin{table}[tbp]
\begin{tabular}{c|ccc}  \hline
Benchmark       & Original & ICLiP   & \textbf{Improve}        \\ \hline
CMath           & 0.467    & 0.867 & \textbf{0.400} \\
MGSM\_zh             & 0.467    & 0.867 & \textbf{0.400} \\
Gaokao2023EN    & 0.467    & 0.867 & \textbf{0.400} \\
CollegeMath     & 0.467    & 0.733 & \textbf{0.266} \\
Minerva Math    & 0.000    & 0.600 & \textbf{0.600} \\
Multi\_LogiEval & 0.602    & 0.733 & \textbf{0.131} \\ \hline
\textbf{AVG} & 0.412 &	0.778 &	\textbf{0.366} \\ \hline
\end{tabular}
\caption{Consistency on 6 models from Llama family}
\label{tb:opensource_consitent}
\end{table}

\subsubsection{Experiments on Open-Source Models}
We select six base models across three released Llama versions (Llama-3-8B, Llama-3-70B, Llama-3.1-8B, Llama-3.1-70B, Llama-3.2-1B, Llama-3.2-3B) along with their corresponding instruction-tuned models, to calculate the capability consistencies on each benchmark. As shown in Table \ref{tb:opensource_consitent}, BOSE significantly enhances the rank correlation of evaluation scores between base and instruct models compared to the original method, achieving an average Kendall's $\tau$ coefficient improvement of 0.366 (+88.9\%).

\begin{figure}[t]
  \includegraphics[width=\columnwidth]{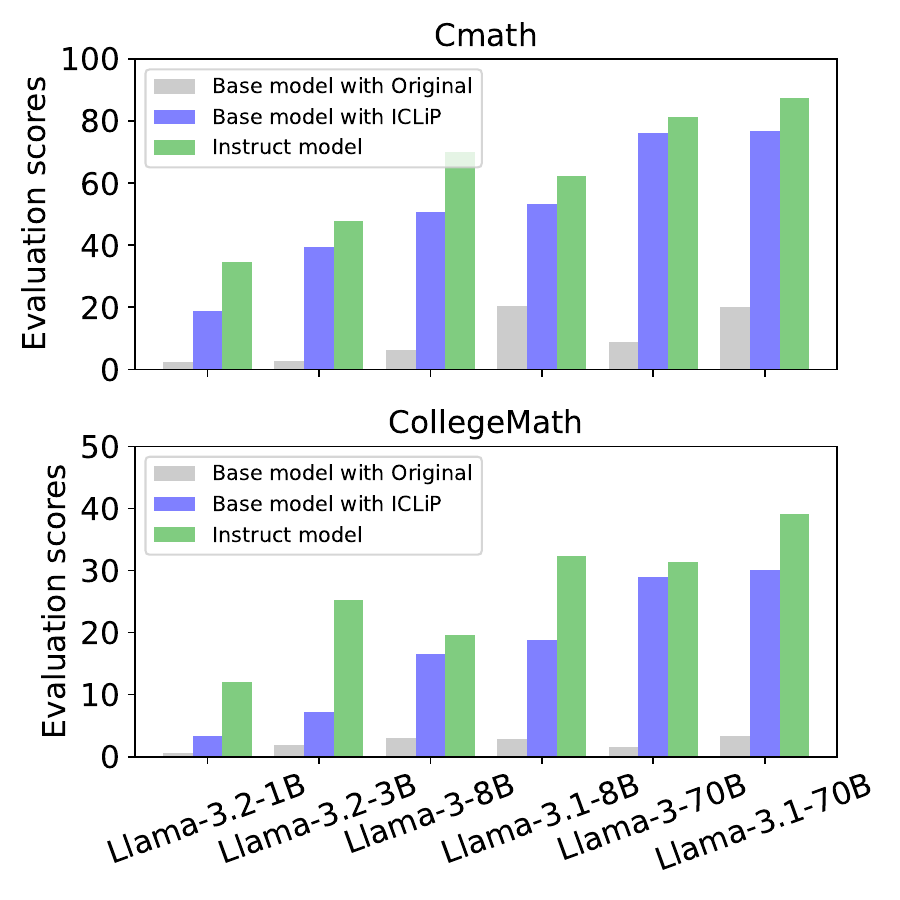}
  \caption{With Llama family models, ICLiP ensures more consistent and comparable evaluation scores between base models and instruct models for Cmath and CollegeMath.}
  \label{fig:llama_consitency}
\end{figure}

Intuitively, we take Cmath and CollegeMath as examples to visualize the evaluation results. As shown in Figure \ref{fig:llama_consitency}, the base and instruct models exhibit stronger trend consistency, providing more coherent capability rankings; moreover, we observe that BOSE enables pre-trained models to achieve comparable performance to their instruction-tuned models, ensures more reliable evaluation results.

\subsubsection{Experiments on Our Pre-trained Models}
Similarly, we conduct another experiment on a series of our pre-trained base models(16 checkpoints in total, with each undergoing supervised fine-tuning to obtain the corresponding instruct model). As shown in Table \ref{tb:pretrain_consitent}, by calculating the rank correlation of the scoring sequences between base and instruct models for each benchmark, we observe an effective improvement in 5 out of 6 benchmarks, with an average Kendall's $\tau$ enhancement of 0.188.

\begin{table}[tbp]
\begin{tabular}{c|ccc}  \hline
Benchmark       & Original & ICLiP    & \textbf{Improve}         \\ \hline
CMath           & -0.070   & 0.367  & \textbf{0.437}  \\ 
MGSM\_zh             & 0.050    & -0.150 & \textbf{-0.200} \\
Gaokao2023EN    & 0.020    & 0.128  & \textbf{0.108}  \\
CollegeMath     & 0.142    & 0.483  & \textbf{0.341}  \\
Minerva Math    & 0.314    & 0.507  & \textbf{0.193}  \\
Multi\_LogiEval & -0.017   & 0.230  & \textbf{0.247}  \\ \hline
\textbf{AVG} & 0.073 &	0.261 &	\textbf{0.188} \\ \hline
\end{tabular}
\caption{Consistency on our pre-trained models}
\label{tb:pretrain_consitent}
\end{table}

\begin{table*}[ht]
\resizebox{0.98\textwidth}{!}
{
\begin{tabular}{c|c|ccccc|c|c} \hline
\multirow{2}{*}{Models}       & \multirow{2}{*}{Methods} & \multicolumn{5}{c|}{Mathematics}              & \multicolumn{1}{c|}{Reasoning}                                                             & \multirow{2}{*}{AVG} \\ 
                              &                          & CMath                & MGSM\_zh                 & Gaokao2023EN         & CollegeMath        & Minerva Math        & Multi\_LogiEval      &                               \\ \hline
\multirow{4}{*}{Llama-3.1-8B}  & instruct\_0shot          & 20.58                & 1.60                 & 1.04                 & 2.91                 & 0.74                 & 0.00                 & 4.48                          \\
                              & instruct\_fewshot        & 52.46*               & 3.60                 & 1.04                 & 0.32                 & 6.99                 & 7.96                 & 12.06                         \\
                              & light-instruction\_0shot    & 32.51                & 16.00                & 9.35                 & 9.17                 & 9.17                 & 18.47                & 15.78                         \\
                              & \textbf{ICLiP}                    & { \textbf{53.37}} & { \textbf{43.60}} & { \textbf{23.38}} & { \textbf{18.86}} & { \textbf{11.03}} & { \textbf{71.03}} & { \textbf{36.88}}          \\  \hline
\multirow{4}{*}{Llama-3.1-70B} & instruct\_0shot          & 17.67                & 11.60                & 5.97                 & 3.26                 & 0.74                 & 0.00                 & 6.54                          \\
                              & instruct\_fewshot        & 67.76                & 2.80                 & 0.26                 & 0.50                 & 6.25                 & 0.05                 & 12.94                         \\
                              & light-instruction\_0shot    & 45.26                & 47.60                & 16.88                & 22.51                & 0.00                 & 21.03                & 25.55                         \\
                              & \textbf{ICLiP}                    & { \textbf{76.05}} & { \textbf{59.60}} & { \textbf{36.62}} & { \textbf{30.77}} & { \textbf{17.28}} & { \textbf{74.96}} & { \textbf{49.21}}          \\  \hline
\multirow{4}{*}{Gemma-2-9B}   & instruct\_0shot          & 17.30                & 10.00                & 2.08                 & 1.31                 & 3.31                 & 0.05                 & 5.68                          \\
                              & instruct\_fewshot        & 68.21*               & 5.20                 & 0.78                 & 0.43                 & 7.72                 & 49.56                & 21.98                         \\
                              & light-instruction\_0shot    & 5.01                 & 26.80                & 14.29                & 26.76                & 0.00                 & 7.86                 & 13.45                         \\
                              & \textbf{ICLiP}                    & { \textbf{69.13}} & { \textbf{55.20}} & { \textbf{33.77}} & { \textbf{33.56}} & { \textbf{16.18}} & { \textbf{61.65}} & { \textbf{44.92}}          \\  \hline
\multirow{4}{*}{Gemma-2-27B}  & instruct\_0shot          & 3.64                 & 8.00                 & 8.57                 & 2.20                 & 1.84                 & 17.94                & 7.03                          \\
                              & instruct\_fewshot        & { \textbf{75.96}} & 3.20                 & 0.78                 & 0.21                 & 8.46                 & 49.34                & 22.99                         \\
                              & light-instruction\_0shot    & 24.41                & 55.20                & 17.14                & 29.30                & 0.00                 & 17.42                & 23.91                         \\
                              & \textbf{ICLiP}                    & 75.87*               & { \textbf{66.40}} & { \textbf{40.52}} & { \textbf{35.11}} & { \textbf{21.32}} & { \textbf{75.68}} & { \textbf{52.48}}         \\ \hline
\multirow{4}{*}{Qwen2.5-7B}   & instruct\_0shot          & 30.69                & 61.60                & { \textbf{53.51}} & 39.62*               & 22.06                & { \textbf{58.54}} & 44.34                         \\
                              & instruct\_fewshot        & 80.87                & 10.80                & 14.29                & 11.86                & { \textbf{37.87}} & 57.67                & 35.56                         \\
                              & light-instruction\_0shot    & 83.97                & 69.60                & 22.34                & 33.65                & 33.65                & 46.08                & 48.22                         \\
                              & \textbf{ICLiP}           & { \textbf{88.07}} & { \textbf{70.80}} & 51.43                & { \textbf{40.45}} & 23.53                & 57.81*               & { \textbf{55.35}}          \\ \hline
\multirow{4}{*}{Qwen2.5-72B}  & instruct\_0shot          & 74.32                & 68.40                & { \textbf{61.30}} & {  \textbf{42.28}} & {  \textbf{36.03}} & 68.89                & 58.54                         \\
                              & instruct\_fewshot        & 85.52*               & 6.40                 & 4.94                 & 12.89                & 34.93                & 71.54                & 36.04                         \\
                              & light-instruction\_0shot    & { \textbf{85.97}} & 79.20                & 40.78                & 38.37                & 0.00                 & 34.08                & 46.40                         \\
                              & \textbf{ICLiP}                    & 83.06                & { \textbf{81.60}} & 55.06                & 40.26                & 29.04                & { \textbf{74.00}} & { \textbf{60.50}}          \\ \hline
\end{tabular}
}
\caption{Ablation study results on 6 open-source models. Scores in \textbf{bold} indicate the highest scores among 4 methods, while scores with ``*'' indicate sub-optimal scores within 1 point gap.}
\label{tb:ablation_study}
\end{table*}

We conclude that, ICLiP ensures more consistent capabilities between base and instruct models, and enables base models to achieve more reliable  scores as well. This advancement provides more meaningful guidance for assessing the performance of pre-trained models and selecting optimal checkpoints for post-training.


\subsection{Ablation Study}
To investigate whether all components of the proposed ICLiP methodology are indispensable, we conduct ablation experiments here. The following variant prompt templates are considered as comparative methods: (1) \textbf{instruct\_0shot}: instruct prompt with 0-shot, also referred to as the original method, (2) \textbf{instruct\_fewshot}: instruct prompt with few-shot, (3) \textbf{light-instruction\_0shot}: light-instruction prompt with 0-shot, (4) \textbf{ICLiP}, our proposed method.
Here, \verb|light-instruction| refers to the prompt template used in our proposed method, while \verb|instruct prompt| is the prompt template typically used for instruct model evaluation. Please refer to Appendix \ref{sec:appendixA} for further implementation details.

\begin{figure}[hp]
  \includegraphics[width=\columnwidth]{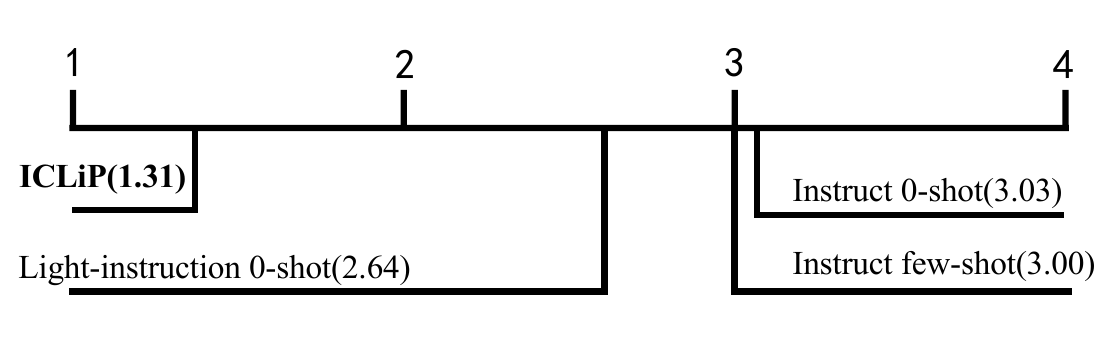}
  \caption{Average ranks of different methods, with a lower rank indicating better performance.}
  \label{fig:method_rank}
\end{figure}

As demonstrated in Table \ref{tb:ablation_study}, ICLiP achieves optimal performance in 28 out of 36 experiment results and suboptimal performance in 2 cases with marginal differences (within 1-point gap). In terms of different LLM families, ICLiP shows dominant superiority for models from Llama and Gemma families, while achieving competitive performance in majority of experimental results with Qwen family. Moreover, the proposed method shows consistently superior average performance across different benchmarks, confirming the effectiveness and generalizability of ICLiP in evaluating base models.

More interestingly, we introduce a statistical analysis of the average rankings of 4 methods (ranking among 4 methods for each experiment, and then averaging across 36 experiments, with 1 representing the best theoretically). As depicted in Figure \ref{fig:method_rank}, ICLiP behaves optimally with an excellent average rank of \verb|1.31|. Additionally, the \verb|light-instruction| format consistently performs better than the instruct prompt, and the few-shot results always outperform the 0-shot results under the same prompt template, highlighting the effectiveness of each component in ICLiP.

To further sourcing the improvements of our proposed method, we conduct two supplementary investigations, and once again proves the validity of the proposed approach. Evaluation details can be found in Appendix \ref{sec:appendixB} .

\section{Conclusion}
In this paper, we propose a systematic evaluation method tailored to base model, named BOSE, specifically designed to enhance the stability of base model evaluations during pre-training and ensure consistency with evaluations of instruct models.
For open-ended tasks, BOSE develops a few-shot prompt template with light-instruction to guide the base model more effectively.
For multi-choice tasks, BOSE innovatively transforms the standard perplexity (ppl) metric into a fill-in-the-blank format. 
These adaptations align better with the inherent characteristics of base models and facilitate more accurate evaluations to reflect the real capabilities of base models. 
Moreover, we are the first to adopt Kendall's rank correlation to quantitatively assess the stability and consistency of base models' evaluation. 
This metric provides a robust and reliable way to compare the evaluation results of the base models with those of corresponding instruct models. 
Extensive experiments demonstrate the effectiveness and superiority of our proposed BOSE, validating its potential to significantly improve the evaluation and provide more meaningful insights into the true capabilities of base models.

\section{Limitations}
Our work aims to address some critical issues in base model evaluation, such as instability during pre-training and lack of consistency with instruct models. There may be some subjective biases in our comparative methods, such as the design of instruction prompt templates. As for the considered models, we use our pre-trained checkpoints and 3 open-source LLM families, and there are more open-source LLMs worth exploring, such as DeepSeek~\cite{deepseekv3}, ChatGLM~\cite{chatglm3}, Mistral~\cite{mistral7b}. At the same time, we mainly conduct experiments on three categories of benchmarks, and some other capability categories (such as code, reading comprehensive, etc.) also need to be considered to provide more comprehensive and enriching experimental results. In addition, our method still has some flaws in some experiments, and we are consistently studying and optimizing the proposed method.

\section*{Acknowledgments}
We sincerely thank the anonymous reviewers for their valuable comments. 

\bibliography{custom}

\clearpage
\appendix

\section{Prompt Templates}
\label{sec:appendixA}
Prompt templates in different methods are illustrated in table \ref{tb:option-ppl} to table \ref{tb:ICLiP}, instruct\_0shot and Option-ppl are regarded as original methods in our experiments for open-ended tasks and multi-choice tasks respectively.

\begin{table}[hp]
\centering
\begin{tabular}{p{7cm}}\hline
\textbf{PROMPT}\\ \\
Question: \texttt{\{question\}} \\
A. \texttt{\{content of optionA\}} \\  
B. \texttt{\{content of optionB\}} \\ 
C. \texttt{\{content of optionC\}} \\ 
D. \texttt{\{content of optionD\}} \\ 
Answer: \uline{A} 
\\ \hline
\end{tabular}
\caption{Prompt of Option-PPL, where the underlined part is used for calculating the Perplexity (PPL).}
\label{tb:option-ppl}
\end{table}

\begin{table}[hp]
\centering
\begin{tabular}{p{7cm}}\hline
\textbf{PROMPT}\\ \\
\{question\} \{\uline{content of optionA}\} \\ \hline
\end{tabular}
\caption{Prompt of Blank-PPL, where the underlined part is used for calculating the Perplexity (PPL).}
\label{tb:blank-ppl}
\end{table}

\begin{table}[hp]
\centering
\begin{tabular}{p{7cm}}\hline
\textbf{PROMPT} \\ \\
You're a \{domain\} expert. Given the following question, please reason step by step and put your final answer within boxed\{\}. \\
\{question\} \\ \hline
\end{tabular}
\caption{Prompt of Instruct\_0shot}
\label{tb:Instruct-0shot}
\end{table}

\begin{table}[hp]
\centering
\begin{tabular}{p{7cm}}\hline
\textbf{PROMPT} \\ \\
You're a \texttt{\{domain\}} expert. Given the following question, please reason step by step and put your final answer within boxed\{\}. \\
\texttt{\{question\_1\}} \\
\texttt{\{answer\_1\}}\\ \\ 
... \\ \\
\texttt{\{question\_k\}} \\ 
\texttt{\{answer\_k\}} \\ \\
\texttt{\{question\}} \\ \hline
\end{tabular}
\caption{Prompt of Instruct\_fewshot}
\label{tb:Instruct-fewshot}
\end{table}

\begin{table}[hp]
\centering
\begin{tabular}{p{7cm}}\hline
\textbf{PROMPT} \\ \\
Problem: \texttt{\{question\}}\\
Solution:  \\ \hline
\end{tabular}
\caption{Prompt of Light-instruction\_0shot}
\label{tb:ICLiP-0shot}
\end{table}

\begin{table}[hp]
\centering
\begin{tabular}{p{7cm}}\hline
\textbf{PROMPT} \\ \\
Problem: \texttt{\{question\_1\}} \\
Solution: \texttt{\{answer\_1\}} \\ \\
... \\ \\
Problem: \texttt{\{question\_k\}} \\
Solution: \texttt{\{answer\_k\}} \\ \\
Problem: \texttt{\{question\}}\\
Solution:  \\ \hline
\end{tabular}
\caption{Prompt of ICLiP}
\label{tb:ICLiP}
\end{table}

\begin{table*}[!htb]
\resizebox{0.98\textwidth}{!}
{
\begin{tabular}{c|cccc|cccc} \hline
\multirow{2}{*}{Benchmark}       &  \multicolumn{4}{c|}{Instruct\_{0shot}}   & \multicolumn{4}{c}{ICLiP}  \\
 & current  &  math\_verify &  upper bound &  ratio &  current  &   math\_verify &  upper bound &  ratio \\ \hline 
CMATH & 20.58 & 16.21 & 22.67 & 0.715 & 53.47 & 52.10 & 53.91 & 0.966 \\
Gaokao2023EN & 1.04 & 0.26 & 4.94 & 0.053 & 23.38 & 23.38 & 24.42 & 0.957 \\
CollegeMath & 2.91 & 9.22 & 9.94 & 0.928 & 18.86 & 23.42 & 24.41 & 0.959 \\ \hline
\end{tabular}
}
\caption{Ablation study results for sourcing the improvement. \textbf{$current$}: results with current evaluation metrics; $math\_verify$ and $upper bound$: results evaluated with Math\_verify and LLM judge respectively; $ratio$: score with math\_verify / score with LLM judge, indicating the extent to which the LLM's predictions follow the format in different methods.}
\label{tb:ablation2}
\end{table*}

\section{Ablation study for sourcing the improvement}
\label{sec:appendixB}
We provide some additional ablation studies to further source the improvements of our proposed method. Specifically, with open-ended tasks in consideration, we take Instruct\_0shot as a comparative method for simplicity and conduct two supplementary investigations:

\textbf{i. Comparable evaluation with math\_verify\footnote{https://github.com/huggingface/Math-Verify}}. We take the newly released mathematical expression evaluation package math\_verify from huggingface to conduct the post-processing and judgment, to identify potential limitations in our current evaluation process.

\textbf{ii. Loose judgement with LLM as upper bound}. To further investigate whether our proposed method effectively ensuring the model's problem-solving capabilities or simply fitting the answer extraction pattern, we employ GPT-4o to assess the correctness of Base model's prediction, where only question (NOT prompt) along with prediction and gold answer are provided for judgment and thus eliminate the influence of answer extraction. These scores are regarded as upper-bound evaluation results, with no constraint imposed on the answer format.

As shown in Table \ref{tb:ablation2}, we observe that our improvement sources from two aspects:

\textbf{Better reflection of true capability}. While evaluation scores slightly differ across different post-processing methods (relatively bigger difference in the CollegeMath), our proposed method consistently outperforms the comparative methods(i.e., Instruct\_0shot).

\textbf{Higher fitness to the answer extraction}. ICLiP exhibits superior alignment with loose scores(upper-bound) compared to both current evaluation process and math-verify judgement, thereby enabling higher tolerance for the post-processing.

The above results further demonstrate the effectiveness of the proposed method.


\balance

\section{Architecture and Pretraining Setup}
\label{sec:arch}

We use a GQA (Grouped Query Attention) architecture based on the standard decoder-only Transformer, comprising an embedding layer, alternating layers of attention mechanisms and feed-forward networks. Positional information is handled using RoPE (Rotary Positional Embedding).
For training parameters, the pre-trained models are initialized with a standard deviation of $0.006$ and optimized using AdamW, with hyperparameters set to $\beta_1 = 0.9$, $\beta_2 = 0.95$, $\epsilon = 1e^{-8}$ and \texttt{weight\_decay} = 0.1. The learning rate follows a WSD (Warm-up, Stabilization, Decay) strategy, where the first $1\%$ of training steps involve linear warm-up. Further architectural and pretraining details are available in Table~\ref{tb:Architecture} and our technical report~\cite{ling2025}.

\begin{table}[!t]
\resizebox{0.48\textwidth}{!}
{
\begin{tabular}{c|cc} \hline
 &  1B &  2B  \\ \hline  
 
vocab size  & 126464 & 126464  \\
layer num & 22 & 22  \\
hidden size & 2048 & 2560  \\
intermediate size & 5632 & 10240  \\ 
attention heads & 32 & 40  \\
key value heads & 4 & 8  \\
\# Para. (B) & 1.487 & 2.724  \\
\# Non-emb. Para. (B) & 0.969 & 2.076  \\ \hline

sequence length & 4096 & 4096  \\
learning rate  & 6.37E-04 & 5.23E-04  \\
batch size & 806 & 1227  \\

\hline
\end{tabular}
}
\caption{The details of architecture and training parameters for pre-trained models in Section \ref{tb:Models}}
\label{tb:Architecture}
\end{table}

\end{document}